\pdfoutput=1
\documentclass[11pt]{article}
\usepackage[]{ACL2023}
\usepackage[T1]{fontenc}
\usepackage[utf8]{inputenc}
\usepackage{microtype}
\usepackage{inconsolata}

\usepackage{amsmath,amsfonts,bm}
\usepackage{calc}
\usepackage{hyperref}
\usepackage{url}
\usepackage{times}
\usepackage{latexsym}
\usepackage{multirow,graphicx}
\usepackage{graphics}
\usepackage{booktabs}
\usepackage{todonotes}
\usepackage{tcolorbox}
\usepackage{fancybox}
\usepackage{caption}
\usepackage{wrapfig}
\usepackage{subcaption}
\graphicspath{ {./images/} }

\newtheorem{researchq}{RQ}
\newenvironment{question}
{\noindent\begin{Sbox}\begin{minipage}{\linewidth-7.5\fboxrule-2\fboxsep-1pt}\begin{researchq}}
{\end{researchq}\end{minipage}\end{Sbox}\doublebox{\TheSbox}}

\title{Exploring Distributional Shifts in Large Language Models for Code Analysis}

\author{Shushan Arakelyan \\
  University of Southern California \\
  \texttt{shushana@usc.edu}\\
  \And
  Rocktim Jyoti Das \\
  IIT, Delhi \& MBZUAI\\
  \texttt{rocktimjyotidas@gmail.com} \\
  \AND
  Yi Mao \\
  Microsoft Azure AI \\
  \texttt{maoyi@microsoft.com}\\
  \And
  Xiang Ren \\
  University of Southern California\\
  \texttt{xiangren@usc.edu}\\
  }

\begin{document}

\maketitle
\vspace{-0.2cm}
\begin{abstract}
We systematically study how three large language models with code capabilities - CodeT5, Codex, and ChatGPT - generalize to out-of-domain data. 
We consider two fundamental applications - code summarization, and code generation.  
We split data into domains following its natural boundaries - by an organization, by a project, and by a module within the software project. 
We establish that samples from each new domain present all the models with a significant challenge of distribution shift. 
We study how established methods adapt models to better generalize to new domains. 
Our experiments show that while multitask learning alone is a reasonable baseline, combining it with few-shot finetuning on examples retrieved from training data can achieve very strong performance. 
Moreover, this solution can outperform direct finetuning for very low-data scenarios.
Finally, we consider variations of this approach to create a more broadly applicable method to adapt to multiple domains at once.
We find that for code generation, a model adapted to multiple domains simultaneously performs on par with those adapted to a single domain\footnote{Code and data for the paper are available at \url{https://github.com/ShushanArakelyan/code_shift/}}. 
\end{abstract}

\vspace{-0.2cm}
\section{Introduction}
\vspace{-0.2cm}
Since the late 2000s, researchers have been reporting poor generalization of statistical learning models to new software systems~\citep{DBLP:journals/ese/Turhan12, DBLP:conf/sigsoft/ZimmermannNGGM09}, a phenomenon that has become important with the rise of large language models (LLMs) for code, such as GitHub Copilot, Amazon CodeWhisperer, Replit, etc. 
Thus, it is crucial to understand when pretrained large language model performance on a private software system will differ from the performance obtained on a benchmark. 
Prior work has studied some aspects of this problem, among others studying generalization from older to newer code, large software projects, and small competition problems, authors, and code representations~\citep{DBLP:conf/acl/NieZLMG22, DBLP:journals/corr/abs-2107-10989, hu2022codes}.

\begin{figure}[h]
 \centering
 \vspace{-0.2cm}
 \includegraphics[trim=0 170 0 40,clip,width=0.45\textwidth]{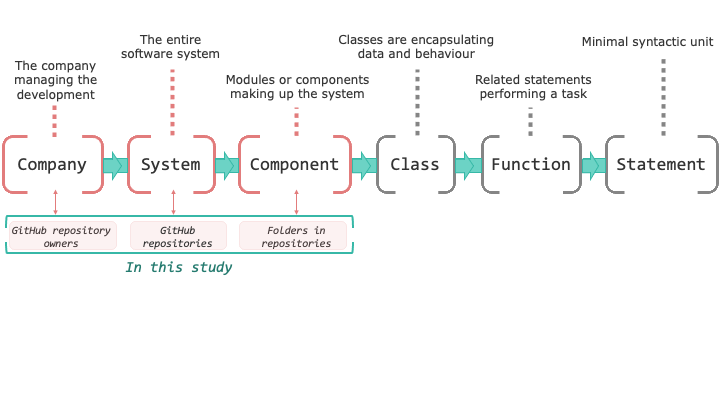}
\caption{Organization of a software system by the granularity of its components}
\vspace{-0.2cm}
\label{fig:motivating}
\end{figure}

However, the challenges of distribution shifts stemming from the hierarchical nature of software data, as depicted in Figure~\ref{fig:motivating}, have not been systematically studied with regard to large language models for code.
Motivated by that, in this work, we probe the generalization capacity of large language models with code capabilities, specifically Codex~\citep{Chen2021EvaluatingLL}, CodeT5~\citep{Wang2021CodeT5IU} and ChatGPT, in code generation and summarization tasks, examining three scenarios: generalization across companies, projects, and project components.
These scenarios are routinely considered for analyzing software systems~\citep{DBLP:journals/infsof/MaLZC12,DBLP:journals/eswa/LiXG09,DBLP:journals/jss/MairKLPSSW00} due to the careful consideration that goes into combining or separating such entities.

First, we want to understand \emph{how models perform on new domains} - if models struggle with out-of-domain generalization, they should be used with caution. 
At the same time, we empirically establish the legitimacy of our definitions for out-of-domain scenarios by demonstrating that these examples present a distributional shift. 
To answer this question, we compare the performance of the models without any additional adaptation with that of the models that have been adapted on limited data from a random domain or from the test domain. 
Adaptation with labeled examples from the test domain is the proxy for model performance if there were no distributional shift. 
We find that all three models suffer from a drop in performance when applied out-of-domain. 
In this experiment, the difference is more pronounced for code summarization, where adapting models with few in-domain examples, on average, leads to an improvement of over 10 BLEU~\citep{papineni-etal-2002-bleu} score points. 

Next, we explore ways to \emph{improve the out-of-domain generalization of large language models with code capabilities}, recognizing that relying on labeled in-domain data for every new domain is impractical. 
Instead, we investigate the use of labeled out-of-domain data and small amounts of \textit{unlabelled} in-domain data to enhance generalization.
We test methods known to be successful in other transfer learning scenarios, such as meta-learning~\citep{DBLP:books/sp/98/ThrunP98,DBLP:journals/air/VilaltaD02} and multitask learning~\citep{DBLP:conf/icml/Caruana96,DBLP:journals/connection/Silver96}.
We also leverage unlabeled in-domain data to retrieve similar labeled examples from an out-of-domain corpus for adapting to the new domain.
We find that while meta-learning and multitask learning do not solve the out-of-domain generalization problem, domain adaptation with retrieved examples is a good technique for low-data domains. In our evaluation on CodeSearchNet dataset we find that models supervised with retrieved examples perform on par, or better, than models that have been adapted using a few samples (e.g., 8 or 16) of in-domain labeled data. 
We are particularly interested in scenarios with an extreme scarcity of labeled data - ranging from a few labeled instances to no labeled data at all. This is due to how new data emerges in software engineering domains - it is not difficult to imagine a new repository, or a new module, with fewer than 32 functions, let alone - 32 labeled functions.

Lastly, \emph{we study if we can make the code models more broadly applicable and retain their generalization capacities}, rather than having to adapt them to every new domain? 
Depending on the approach to model adaptation (e.g. weight update vs in-context demonstrations) we vary the set of retrieved examples for each new domain, or for each test input individually. We compare performance obtained this way with that of the models that are adapted simultaneously to multiple domains (or instances, correspondingly). 
We find that Codex is very sensitive to these changes, so it is best to retrieve similar instances for each test data point.
On the other hand, CodeT5 has a minor drop in code summarization and a negligible drop in code generation. 
This makes it feasible to adapt and apply CodeT5 to multiple domains simultaneously with minimal tradeoff, eliminating the need to store separate copies of the model for each domain.

\vspace{-0.2cm}
\section{Background}
\vspace{-0.2cm}
The shifts in underlying semantics between the training and evaluation data can be one of the most impacting factors for deteriorating performance at test time.
Prior work in code analysis has mainly focused on \textit{cross-project} shifts, i.e. training and evaluating the model on disjunct sets of code projects. 
Additionally, the studies were mainly conducted in the context of \textit{traditional machine learning methods}, such as linear classifiers, support vector machines, and later, LSTMs \citep{DBLP:conf/sigsoft/ZimmermannNGGM09, DBLP:journals/ese/Turhan12, DBLP:conf/itasec/AngioniDPB22}. 

More recent works consider shifts caused by different authors of the code, the timeline of the project, distributions of code tokens, etc~\citep{DBLP:journals/corr/abs-2107-10989,hu2022codes,DBLP:conf/acl/NieZLMG22}. 
However, the abilities of large language models under distribution shift are still under-explored.
We conduct a comprehensive empirical analysis to probe the large language models' capabilities in handling three different granularity of distribution shifts (company, domain, module) when different training and adaptation methods are used. 
In addition to directly fine-tuning vanilla LLMs, we experiment with enhancing pretrained models using the methods described below. 
\vspace{-0.2cm}
\paragraph{Meta-Learning and Multi-task Learning.} In our work, we experiment with both Meta-Learning and Multi-task learning to get better initialization for few-shot performance on the downstream task. For meta-learning, we employ Model-agnostic Meta-Learning (MaML)~\citep{Finn2017ModelAgnosticMF} which is a gradient-based method. It is a conceptually simple and model-agnostic algorithm that has been shown to outperform existing approaches in several tasks. Multi-task Learning (MTL) aims to learn a shared and generalized representation by jointly training on several tasks. We adopt the simplest approach to multi-task learning by jointly finetuning a shared language model on multiple tasks. 
\vspace{-0.2cm}
\paragraph{Parameter Efficient Methods.} Parameter-efficient methods have been shown to obtain performance comparable to finetuning all model parameters with finetuning only a tiny fraction of model parameters. In our work, we experiment with Low-Rank Adaptation~(LoRA)~\citep{Hu2021LoRALA}, which is a low-rank update method.
\vspace{-0.2cm}
\paragraph{In-Context Learning.}
GPT-3~\citep{Brown2020LanguageMA} demonstrated the ability of large language models to perform few-shot predictions, where the model is given a description of the task in natural language with few examples. In our work, we conduct experiments on in-context learning on Codex.
\vspace{-0.2cm}
\paragraph{Retrieval Based Example Selection.} It has been shown in~\citet{Liu2021WhatMG} that in-context examples selected following a strategy may serve as more informative input to unleash GPT3's extensive knowledge. Inspired by this, we leverage a similarity-based retrieval for domain adaptation.

\vspace{-0.2cm}
\section{Problem setting}
\vspace{-0.2cm}
\begin{figure}[h]
 \centering
 \includegraphics[trim=0 180 430 0,clip,width=0.4\textwidth]{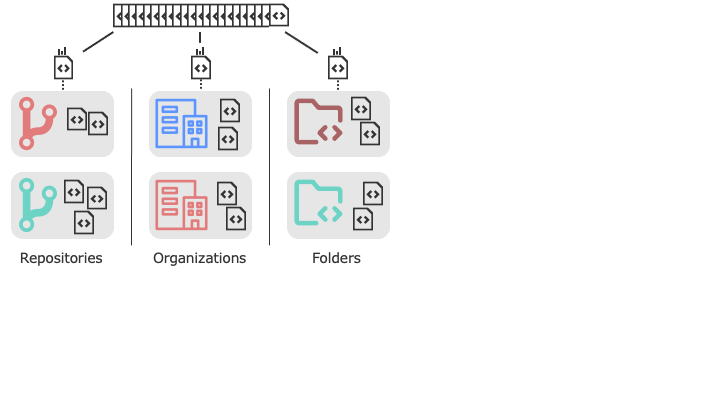}
\caption{We group the functions from CodeSearchNet by repos, orgs, and folders they belong to.}
\vspace{-0.2cm}
\label{fig:data}
\end{figure}
We study scenario where users seek to integrate a large language model, such as Codex or CodeT5, into their software project. The primary focus of this study is to gain a deeper understanding of the performance characteristics exhibited by these models, particularly when confronted with source code originating from an unseen organization, an unseen project, or specific project components that have not been previously encountered. 

For every code data point in the dataset, we have information about the \textit{organization}, \textit{project}, and the \textit{module} within the project that the data point comes from. 
Based on this information, we can group data points into sets, and end up with three \textit{sets of sets}, as illustrated in Figure~\ref{fig:data}. 
For example, the middle set in the figure contains multiple sets of data points. 
Each of those sets corresponds to a unique organization to which all data points within it belong. 
In other words, all data points within a set belong to the same domain. Appendix, Section~\ref{app:data-viz} contains additional analysis on splitting the data points in this manner.
For simplicity, we refer to a set of examples from the same domain as $\tau^i$. 
We refer to splits of such a set into train, development, or test sections as $\tau^{i}_{train}$, $\tau^{i}_{dev}$, and $\tau^{i}_{test}$. 

\vspace{-0.2cm}
\subsection{Data} \label{sec:data}
\vspace{-0.2cm}
We use CodeSearchNet~\citep{Husain2019CodeSearchNetCE} dataset\footnote{Since the training data of Codex models is undisclosed, we cannot be sure that it did not include CodeSearchNet. Nevertheless, we see a performance difference for ID and OOD experiments.}, in particular, the partition containing JavaScript language. 
We refer to the train section of the dataset as $X_{train}$, and the development and test sections as $X_{test}$. 

\begin{table}
\centering
\resizebox{0.4\textwidth}{!}{
\begin{tabular}{lrrr}
\toprule
 & $\tau \subset X_{train}$ (total) & $\tau \subset X_{train} (|\tau | \geq 96$)  & $\tau \subset X_{test} (|\tau| \geq 96$)\\
 \midrule
org. & 9737 & 195 & 8 \\
repos. & 15858 & 147 & 15 \\
fold. & 25268 & 100 & 10 \\
\bottomrule
\end{tabular}}
\caption{Domains in CodeSearchNet dataset. Left column: training set. Middle column: number of domains of each kind in $X_{train}$ with > 96 samples. Right column: number of domains in $X_{test}$ with > 96 samples.}
\vspace{-0.2cm}
\label{tab:dataset}
\end{table} 

We want to keep all of the domains in $X_{test}$ unseen, and for that reason, we remove any domain from $X_{test}$ that also appears in $X_{train}$. This can happen because CodeSearchNet dataset is split into partitions by projects, so the same organizations can appear in different splits. This way, any domain coming from $X_{test}$ is, by our definition, out-of-domain for any model trained on $X_{train}$. 
We further split each domain $\tau^i \subset X_{test}$ into $\tau^i_{train}$, $\tau^i_{dev}$ and $\tau^i_{test}$. 
The evaluation is performed on $\tau^i_{test}$. $\tau^i_{train}$ and $\tau^i_{dev}$ are used to obtain a proxy for the upper-bound performance of the model if the domain $\tau^i$ was seen during training, i.e. if there is no distribution shift for $\tau^i_{test}$.

\paragraph{Preprocessing}
We use the ``path'' field of the data point to determine each code snippet's organization, repository, and lowest-level folder.
Using 5 different random seeds, we divide a domain into $\tau^i_{train}$, $\tau^i_{dev}$, and $\tau^i_{test}$. 
We aim to have at least 32 samples each in $\tau^i_{test}$ and $\tau^i_{dev}$, and up to 32 samples for $\tau^i_{train}$. 
Thus, from $X_{test}$ we filter any domain that has less than 96 samples in total. 
Final dataset statistics are presented in Table~\ref{tab:dataset}. 

\begin{table*}[t]
\centering
\resizebox{0.9\textwidth}{!}{%
\begin{tabular}{lrrrrrrrrr}
\toprule
\multicolumn{1}{c}{
    \multirow{2}{*}{\textbf{Code summarization}}
} &
\multicolumn{3}{c}{\textbf{folder}} & 
\multicolumn{3}{c}{\textbf{repo}} & 
\multicolumn{3}{c}{\textbf{org}} \\
\cmidrule(lr){2-4}
\cmidrule(lr){5-7}
\cmidrule(lr){8-10}
\multicolumn{1}{c}{} & 
\multicolumn{1}{c}{\textbf{8-shot}} & 
\multicolumn{1}{c}{\textbf{16-shot}} & 
\multicolumn{1}{c}{\textbf{32-shot}} & 
\multicolumn{1}{c}{\textbf{8-shot}} & 
\multicolumn{1}{c}{\textbf{16-shot}} & 
\multicolumn{1}{c}{\textbf{32-shot}} & 
\multicolumn{1}{c}{\textbf{8-shot}} & 
\multicolumn{1}{c}{\textbf{16-shot}} & 
\multicolumn{1}{c}{\textbf{32-shot}} \\
\midrule
CodeT5 FT ID 
& 14.39 & 16.06 & 18.31
& 12.68 & 14.73 & 16.82
& 13.14 & 16.35 & 17.65  \\
CodeT5 LoRA ID 
& 16.57 & 19.07 & 20.93
& 15.22 & 17.14 & 21.20
& 15.61 & 18.56 & 20.87  \\
CodeT5 FT random
& 3.58 & 4.30 &  5.02
& 4.35 & 4.70 & 5.79
& 4.53 & 5.47 & 6.27  \\
CodeT5 LoRA random 
& 3.69 & 4.37 & 4.92
& 4.70 & 5.56 & 5.92
& 5.27 & 5.53 &  6.26 \\
\bottomrule          
\end{tabular}
}
\caption{Model performance for code summarization on in-domain (\textbf{ID}) vs out-of-domain (\textbf{random}) test data. Reported metric is BLEU (higher is better). 
}
\vspace{-0.2cm}
\label{tab:code-to-text}
\end{table*}

\begin{table*}[t]
\centering
\resizebox{0.9\textwidth}{!}{%
\begin{tabular}{lrrrrrrrrr}
\toprule
\multicolumn{1}{c}{
    \multirow{2}{*}{\textbf{Code generation}}
} &
\multicolumn{3}{c}{\textbf{folder}} & 
\multicolumn{3}{c}{\textbf{repo}} & 
\multicolumn{3}{c}{\textbf{org}} \\
\cmidrule(lr){2-4}
\cmidrule(lr){5-7}
\cmidrule(lr){8-10}
\multicolumn{1}{c}{} & 
\multicolumn{1}{c}{\textbf{8-shot}} & 
\multicolumn{1}{c}{\textbf{16-shot}} & 
\multicolumn{1}{c}{\textbf{32-shot}} & 
\multicolumn{1}{c}{\textbf{8-shot}} & 
\multicolumn{1}{c}{\textbf{16-shot}} & 
\multicolumn{1}{c}{\textbf{32-shot}} & 
\multicolumn{1}{c}{\textbf{8-shot}} & 
\multicolumn{1}{c}{\textbf{16-shot}} & 
\multicolumn{1}{c}{\textbf{32-shot}} \\
\midrule
CodeT5 FT ID 
& 14.67 & 15.22 & 16.13
& 16.15 & 17.42 & 18.62
& 14.54 & 15.34 & 16.43  \\
CodeT5 LoRA ID 
& 14.14 & 15.06 & 16.36
& 16.23 & 17.45 & 18.96
& 14.17 & 15.30 & 16.62 \\
CodeT5 FT random 
& 15.23 & 14.94 & 15.15
& 14.19 & 14.14 & 14.67
& 13.39 & 13.43 & 14.44 \\
CodeT5 LoRA random 
& 14.45 & 14.29 & 15.37
& 14.29 & 13.74 & 15.04
& 13.76 & 13.85 & 14.81\\
\bottomrule          
\end{tabular}
}
\caption{Model performance for code generation on in-domain (\textbf{ID}) vs out-of-domain (\textbf{random}) test data. Reported metric is CodeBLEU (higher is better).
}
\vspace{-0.2cm}
\label{tab:text-to-code}
\end{table*}

\begin{table}[]
\centering
\resizebox{0.48\textwidth}{!}{%
\begin{tabular}{llrrr}
\toprule 
\textbf{Code summarization} & &
\textbf{folder} & \textbf{repo} & \textbf{org} \\
\midrule
\multirow{3}{*}{Codex} & instr. only \textit{(0-shot)} &  
1.55 & 1.52 & 1.61 \\
& ICL random \textit{(8-shot)} & 7.17 & 6.84 & 6.73 
\\
& ICL ID \textit{ (8-shot)} & 20.34 & 19.00 & 20.72 \\
 \midrule
\multirow{3}{*}{ChatGPT}& instr. only \textit{(0-shot)} & 5.74 & 5.48 & 4.63 \\
 & ICL random \textit{(8-shot)} &  5.47 & 6.58 & 6.48 \\
 & ICL ID \textit{(8-shot)} & 7.47 & 9.15 & 7.54 \\
 \midrule
\textbf{Code generation} & &
\textbf{folder} & \textbf{repo} & \textbf{org} \\
\midrule
\multirow{3}{*}{Codex} & instr. only \textit{(0-shot)} &  5.49 & 5.72 & 5.77 \\
& ICL random \textit{(8-shot)} & 16.82 & 17.47 & 16.82 \\
& ICL ID \textit{ (8-shot)} & 25.73	& 24.64 & 23.87 \\
\midrule
\multirow{3}{*}{ChatGPT}& instr. only \textit{(0-shot)}  & 8.45 & 8.39 & 8.04 \\
& ICL random \textit{(8-shot)} & 12.95 & 13.19 & 12.70 \\
& ICL ID \textit{(8-shot)} & 15.17 & 15.81 & 15.55 \\
\bottomrule
\end{tabular}
}
\caption{Codex and ChatGPT performance for code summarization and generation tasks. Models are evaluated 0-shot, as well as using ICL with in-domain (\textbf{ID}) and out-of-domain (\textbf{random}) data. Reported metric is BLEU for code summarization (higher is better), and CodeBLEU for code generation (higher is better)}
\vspace{-0.6cm}
\label{tab:copilot-table}
\end{table}

\vspace{-0.2cm}
\subsection{Applications and Metrics}
\vspace{-0.2cm}
We study two generation applications: code summarization and code generation. 
\textbf{Code summarization} aims to summarize a code snippet into a natural language description. The code snippet in CodeSearchNet dataset is a function, while the natural language description is the docstring of that function. This task is evaluated with BLEU-4~\citep{papineni-etal-2002-bleu} metric. \textbf{Code generation} generates the function given a natural language description of the code. We follow prior work and use CodeBLEU~\citep{Ren2020CodeBLEUAM} for evaluating generated code. We added our own JavaScript keywords (the full list is in Appendix, Section~\ref{app:js-kwords}) to an existing CodeBLEU implementation. However, recently it has been shown that CodeBLEU scores can disagree with human judgment scores~\cite{DBLP:journals/corr/abs-2208-03133}. Motivated by these findings we additionally evaluate code generation models with chrF~\citep{DBLP:conf/wmt/Popovic15}, RougeL~\citep{lin2004rouge} and CodeBERTScore~\citep{Zhou2023CodeBERTScoreEC} metrics. These metrics are in agreement in our experiments, so we report the results for them in Appendix, Section~\ref{app:metrics}. 
\vspace{-0.2cm}
\subsection{Models}
\vspace{-0.2cm}
We experiment with three large language models: (1) CodeT5~\citep{Wang2021CodeT5IU}, which is an encoder-decoder model based on T5~\citep{Raffel2019ExploringTL}, (2) Codex~\citep{Chen2021EvaluatingLL}, which is a decoder only model based on GPT-3~\citep{Brown2020LanguageMA} and (3) ChatGPT (\textit{gpt-3.5-turbo}) which is the chat optimized version of InstructGPT~\citep{Ouyang2022TrainingLM} which is fine-tuned with Reinforcement Learning with Human Feedback(RLHF)~\citep{Christiano2017DeepRL}. 
The models vary in size: CodeT5 utilizes the T5-large architecture with 700 million parameters, while the Codex model employs the GPT-3 architecture with over 100 billion parameters. Although the architecture of ChatGPT has not been disclosed, it is presumed to have billions of parameters.
A more detailed discussion of these models is provided in the Appendix, Section~\ref{app:model}.
\vspace{-0.2cm}
\section{Analysis}
\vspace{-0.2cm}
In this section, we formulate the research questions that we aim to answer and give a more detailed description of the setups that we have used for analyzing and answering each question. 

\vspace{0.2cm}
\begin{question}
How do code models perform on new domains?
\end{question}

We test the models' capacity for generalization to new domains by comparing the performance of the models that have been adapted to the new domain using few-shot instances of in-domain data (ID) vs those that only encountered out-of-domain (OOD) data. 
For CodeT5, few-shot domain adaptation data is used to update the model weights, whereas for Codex, it is included as demonstrations in the prompt to the model. 
\vspace{-0.2cm}
\subsection*{CodeT5}
\vspace{-0.2cm}
For adaptation techniques for the CodeT5 model, we experiment with using a different number of supervision examples - 8, 16, or 32.

The first adaptation method we use is full model \textbf{fine-tuning} (\textbf{FT}). Information on the hyperparameters for this and all other methods is available in Appendix, Section~\ref{app:hyperparams}. Besides FT, we also experiment with a parameter-efficient fine-tuning method - \textbf{Low-Rank Adaptation} (\textbf{LoRA})~\citep{Hu2021LoRALA}. This method adds trainable pairs of rank decomposition matrices in parallel to existing weight matrices, thus enabling parameter-efficient adaptation to new domains without forgetting. 
\vspace{-0.2cm}
\subsection*{Codex and ChatGPT} 
\vspace{-0.2cm}
For GPT models, we do not perform weight updates. 
Very large models have been shown to be capable to generalize to unseen tasks with just an instruction. 
Thus, we evaluate these models with just the task instruction, for example, "Summarize following JavaScript code", and input (i.e. \textbf{instruction only}).
Models can be sensitive to the wording of the instructions, so we use a number of different instruction variations for each application and average the results. The full list of instruction variations that we have used with Codex and ChatGPT models is presented in Appendix, Section~\ref{app:instructions}. 

Moreover, larger models have been shown to ``learn'' from demonstration examples that are provided as part of their input, even though this process does not involve any weight updates. 
This phenomenon is known as in-context learning (\textbf{ICL}), which is what we use for domain adaptation for GPT models.
Due to the limit on the size of the input to the models (4096 tokens), we use as many demonstrations as would fit, including up to 8 demonstrations with each test example. And since the models can also be sensitive to the order of examples, we shuffle the order of the demonstrations 5 times and average the results. 

\vspace{-0.2cm}
\subsection*{Finding: Models struggle on new domains}
Tables~\ref{tab:code-to-text} and~\ref{tab:text-to-code} demonstrate the performance obtained by CodeT5, and Table~\ref{tab:copilot-table} shows performance for Codex and ChatGPT. Additional results for other code generation metrics, such as chrF, RougeL, and CodeBERTScore are available in Appendix, Section~\ref{app:metrics}.
We see that the performance degrades for models that have not encountered in-domain examples vs those that have, i.e. models struggle with out-of-domain generalization. For example, CodeT5 model on code summarization in most scenarios gains about 200\% relative improvement after updating the model with few-shot in-domain data.

While there is a difference in performance for CodeT5 model on code generation ID and OOD, the performance difference is next to negligible. 
We hypothesize that this can be due to the fact that code generation is a more challenging task for a large language model, and so the effect of distribution shift is less noticeable. 
This observation becomes evident when examining Table~\ref{tab:text-to-code}, which demonstrates that the smaller model, CodeT5, exhibits lower performance compared to larger models such as Codex. 
Thus, for CodeT5 adding in-domain data results in a smaller gain. On the other side, for Codex, the addition of the in-domain data results in up to 50\% relative improvement.

From Table~\ref{tab:copilot-table}, it is evident that while ChatGPT outperforms Codex in 0-shot setting, we don't see as large of an improvement with the addition of in-context examples, whether in-domain or out-of-domain. Upon closer inspection of model outputs, we notice that this is due to the specifics of the ChatGPT model, which errs on the side of caution, refusing to provide any answer when presented with a vague or noisy input. This results in 0 scores for those entries, lowering the overall model performance and smoothing the effect of in-domain demonstrations. Due to this characteristic of ChatGPT model, having established that it is affected by distributional shifts same as other models in this study, we do not perform further comparisons with it in the rest of the paper.

\begin{figure}[t]
     \centering
     \begin{subfigure}[b]{0.4\textwidth}
         \centering
         \includegraphics[trim=70 130 80 20,clip,width=\textwidth]{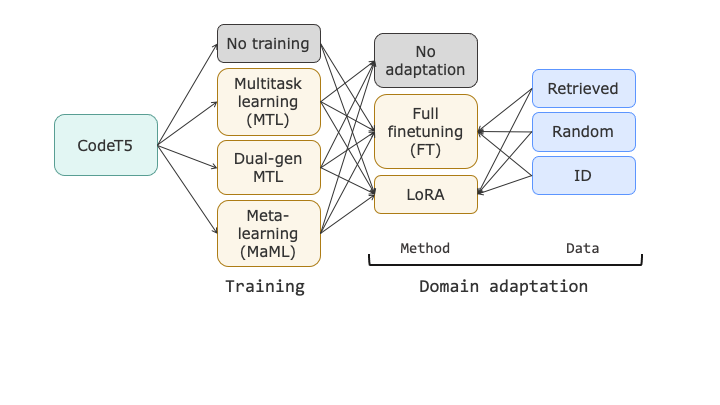}
        \caption{CodeT5}
        \label{fig:methods-codet5}
     \end{subfigure}
     \hfill
     \begin{subfigure}[b]{0.4\textwidth}
         \centering
         \includegraphics[trim=90 170 90 70,clip,width=\textwidth]{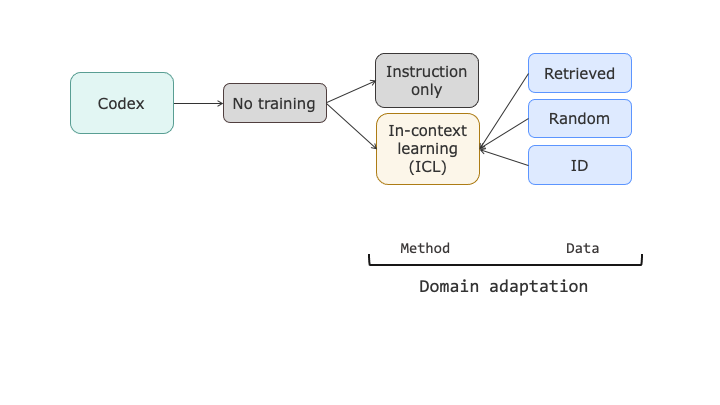}
         \caption{Codex}
         \vspace{-0.2cm}
        \label{fig:methods-codex}
     \end{subfigure}
        \caption{For the CodeT5 model we use different methods for training and domain adaptation. We evaluate both in scenarios with different data sources during the domain adaptation stage.}
        \vspace{-0.2cm}
    \label{fig:methods}
\end{figure}

\begin{figure*}[h]
     \centering
     \begin{subfigure}[b]{\textwidth}
     \begin{subfigure}[b]{0.5\textwidth}
         \centering
         \includegraphics[trim=55 15 60 5,clip,width=\textwidth]{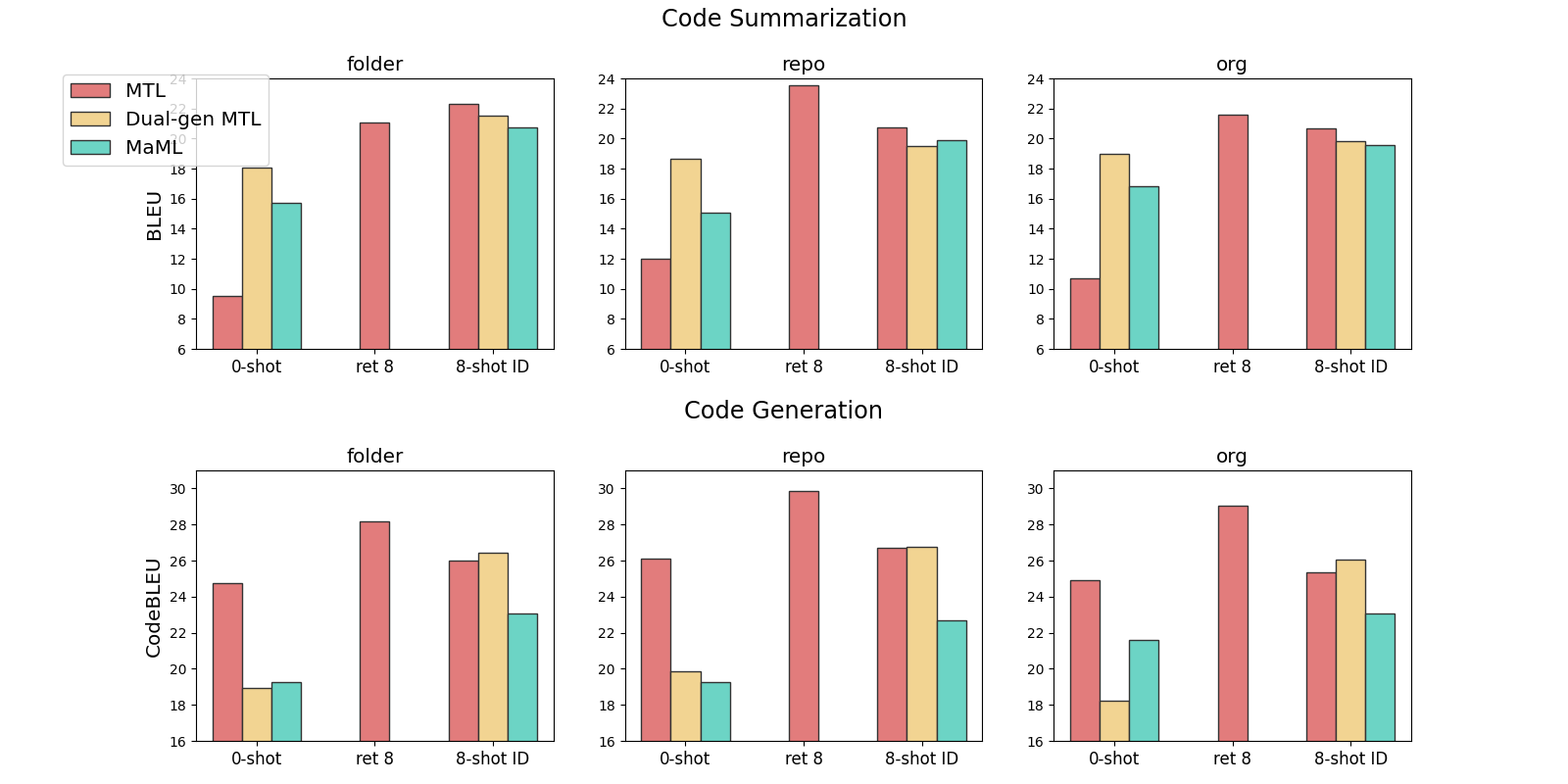}
        \caption{CodeT5, trained and evaluated on CodeSearchNet}
        \label{fig:result2-codet5}
     \end{subfigure}
     \begin{subfigure}[b]{0.5\textwidth}
         \centering
         \includegraphics[trim=55 15 60 5,clip,width=\textwidth]{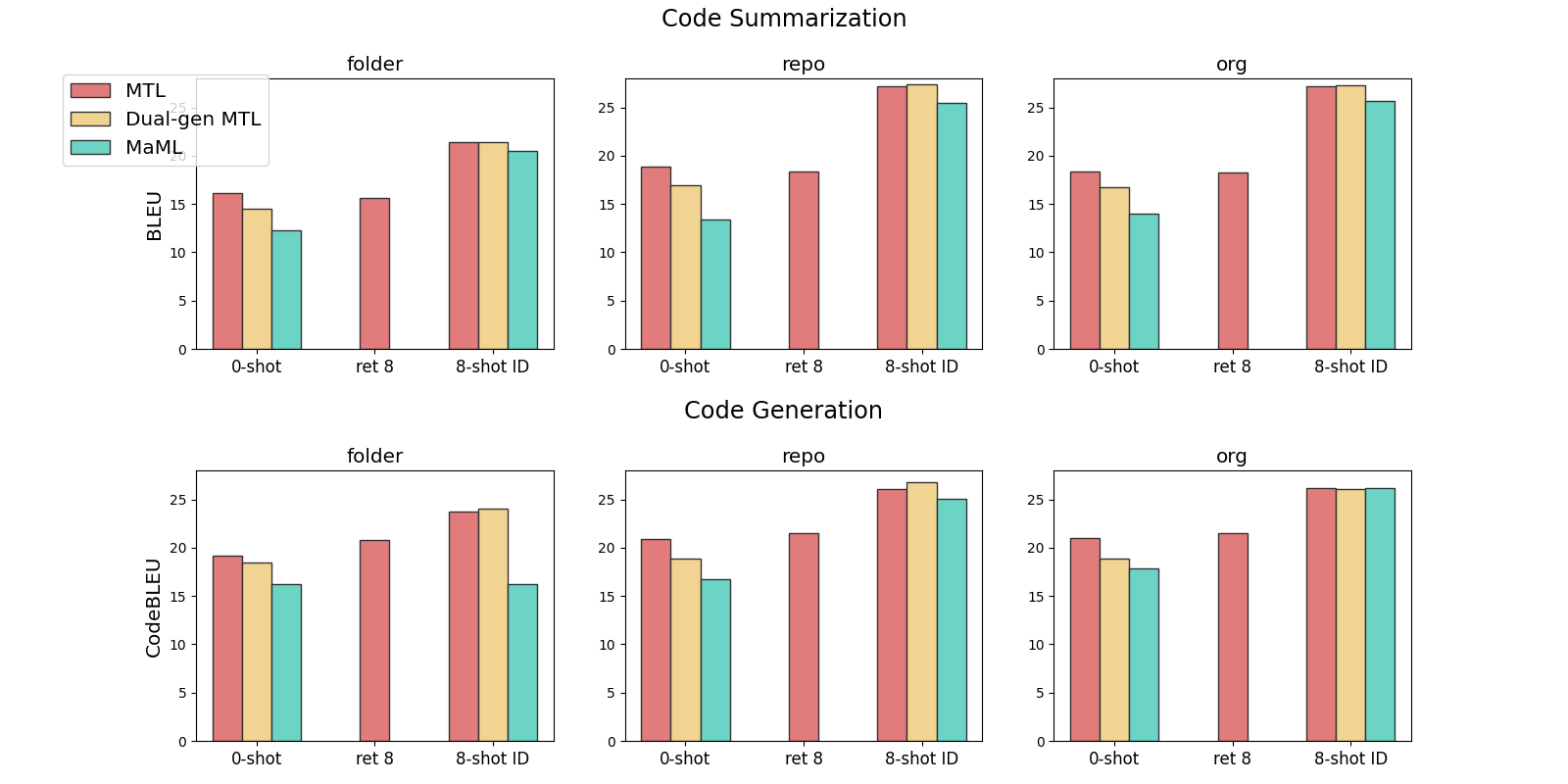}
        \caption{CodeT5, trained on CodeSearchNet, evaluated on The Vault}
        \label{fig:result2-codet5-vault}
     \end{subfigure}
         
     \end{subfigure}
     \hfill
     \begin{subfigure}[b]{0.85\textwidth}
         \centering
         \includegraphics[trim=30 0 50 0,clip,width=\textwidth]{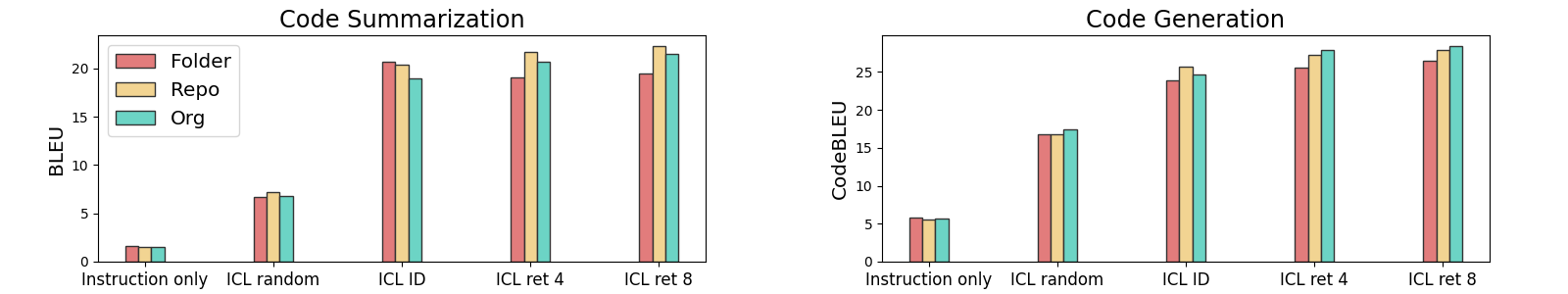}
         \caption{Codex}
         \vspace{-0.2cm}
        \label{fig:result2-codex}
     \end{subfigure}
        \caption{Models with ID and retrieved downstream adaptations. }
        \vspace{-0.2cm}
\end{figure*}

\begin{figure*}[h]
     \centering
     \includegraphics[trim=0 0 0 0,clip,width=\textwidth]{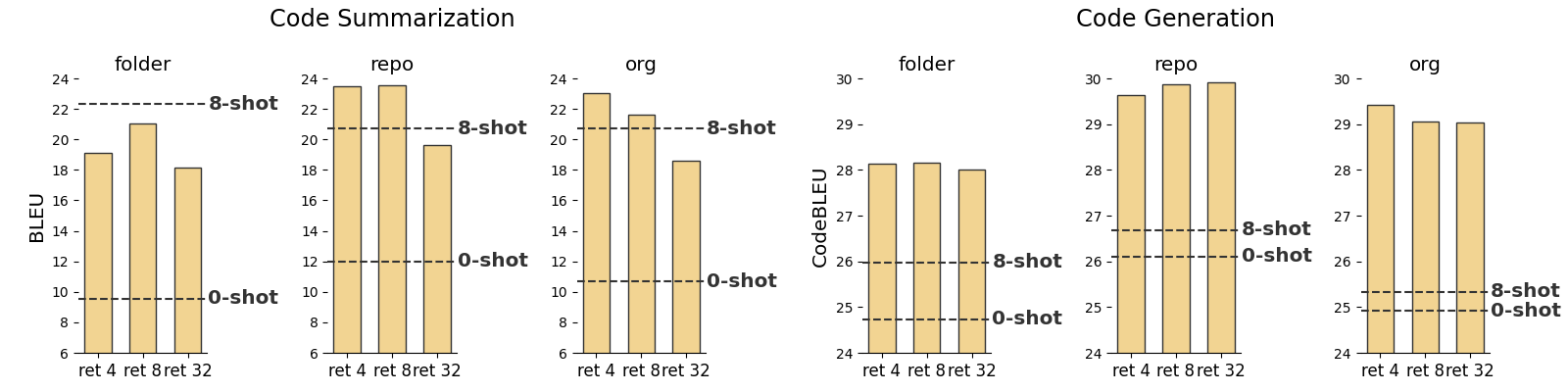}
     \caption{CodeT5 model finetuned with retrieved supervision using different number of retrieved examples per test sample. Scores reported are BLEU for code summarization and CodeBLEU for code generation. CodeT5 MTL model performances in zero-shot, and 8-shot (ID) scenarios are shown with dotted lines for reference.}
     \vspace{-0.2cm}
     \label{fig:result2-var-ret}
\end{figure*}

\begin{question}
How to get better out-of-domain generalization?
\end{question}
We have seen that models for code performed significantly better after being adapted for new domains using in-domain data. 
However, there are many reasons why adapting to every new domain with the help of labeled examples might be impractical. 
Thus, we consider some alternative approaches, that would not require labeled data but can hopefully close the performance gap partially or fully.
Figure~\ref{fig:methods} shows an overview.
\vspace{-0.2cm}
\subsubsection*{CodeT5}
\vspace{-0.2cm}
To answer RQ1, we start from a pre-trained checkpoint of the model and experiment with different approaches for domain adaptation. To answer the current question, we additionally consider different methods to use before the domain adaptation stage, particularly, multi-task learning and meta-learning. The resulting setups are illustrated in Figure~\ref{fig:methods-codet5}.

\vspace{-0.2cm}
\paragraph{Multitask learning (MTL)}
MTL method trains a single model on all the domains simultaneously.
For code summarization, we use the model checkpoint that has been provided by the authors of CodeT5, which is fine-tuned on the training portion of CodeSearchNet. 
For code generation,  we perform our own training since there was no JavaScript checkpoint shared by CodeT5 authors.

\vspace{-0.2cm}
\paragraph{Dual-gen MTL}
In addition to MTL, we experiment with a multitask model that has been trained on both code generation and code summarization simultaneously. We refer to this model as ``dual-gen'' MTL, following the authors of CodeT5. 
We prepend the inputs to the model with a generation or summarization instruction for each instance. 

\vspace{-0.2cm}
\paragraph{Model-Agnostic Meta Learning} 
For model-agnostic meta-learning or MaML~\citep{Finn2017ModelAgnosticMF}, we filter the domains in $X_{train}$ set, only leaving those that have at least 96 samples (see the middle column of Table~\ref{tab:dataset}). This is to ensure that each domain contains disjoint sets of adequate size for both training and meta-training. 

\vspace{-0.2cm}
\paragraph{Stratified Example Retrieval for Supervision}
In addition to the strategies above, we experiment with a domain adaptation method that does not require in-domain labeled data for supervision. 
We use a similarity metric on embeddings obtained from the pre-trained CodeT5 model checkpoint to \textbf{retrieve} $k$ most similar examples for every example in $\tau_{test}$ from $X_{train}$. 
We set $k$ to 4, 8, or 32, and since $|\tau_{test}|=32$ the combined size of the set would be 128, 256, or 1024. 
Finally, we remove any duplicates. 
We refer to this set as $\tau_{ret}$. 

For similarity metric, we experiment with cosine similarity, as well as a more recent approach - IsoScore~\citep{DBLP:conf/acl/RudmanGRE22}. In our experiments, we find that cosine similarity performs better overall, so the results reported in the paper are using cosine similarity. Additional results using IsoScore metric are reported in Appendix Section~\ref{app:similarity-isoscore}.
\vspace{-0.2cm}
\paragraph{Challenge Scenario}
In addition to using test data from CodeSearchNet dataset, in an attempt to make the evaluation more realistic, we experiment with a setting where the out-of-domain data comes from \textit{a different dataset}. Here we use the test split of The Vault dataset \citep{DBLP:journals/corr/abs-2305-06156}, which we have processed in the same manner as described in Section~\ref{sec:data}. The details of the processing for the Vault dataset are provided in Appendix Section~\ref{app:vault}. 

\vspace{-0.2cm}
\subsubsection*{Codex}
\vspace{-0.2cm}
\paragraph{Stratified Example Retrieval for Demonstrations}
Similarly to the strategy for CodeT5, for Codex we employ in-context learning with retrieved demonstration examples. 
For each test query, instead of using random sets of in-domain or out-of-domain demonstrations, we use 4 or 8 of the query's most similar samples from $X_{train}$ as demonstrations. 
This case is referred to as \textbf{ICL ret}.

\begin{table*}[h]
\centering
\resizebox{0.9\textwidth}{!}{\begin{tabular}{lrrrrrr}
\toprule
 & \multicolumn{3}{c}{\begin{tabular}[c]{@{}c@{}}Code Summarization\\ 
 BLEU / $\Delta$ BLEU\end{tabular}} & \multicolumn{3}{@{\hskip 0.6cm}c}{\begin{tabular}[c]{@{}c@{}}Code Generation\\ CodeBLEU / $\Delta$ CodeBLEU\end{tabular}} \\
 \cmidrule(lr){2-4}
 \cmidrule(lr){5-7}
 & \multicolumn{1}{c}{org} & \multicolumn{1}{c}{repo} & \multicolumn{1}{c}{folder} & \multicolumn{1}{c}{org} & \multicolumn{1}{c}{repo} & \multicolumn{1}{c}{folder} \\
\midrule
FT: combined 4 
& 18.74 / -4.74 & 18.59 / -4.47 & 18.06 / -1.06 & 29.46 / -0.19 & 29.41 / -0.01 & 26.60 / -1.53 \\
FT: combined 8 
& 18.46 / -5.07 & 18.58 / -3.03 & 17.57 / -3.48 & 29.13 / -0.73 & 28.83 / -0.22 & 27.23 / -0.92 \\
FT: combined 32 
& 17.35 / -2.31 & 17.63 / -0.94 & 15.57 / -2.56 & 26.28 / -3.63 & 25.01 / -4.02 & 25.14 / -2.88 \\
\midrule
ICL: 4 from $\tau_{ret}$
& 14.66 / -7.04 & 12.68 / -7.95 & 12.10 / -6.96 & 20.52 / -6.73 & 20.06 / -7.78 & 19.39 / -6.21 \\
ICL: 8 from $\tau_{ret}$
& 13.77 / -8.53 & 12.96 / -8.52 & 12.26 / -7.17 & 20.81 / -7.05 & 20.23 / -8.16 & 19.48 / -7.00\\
\bottomrule
\end{tabular}}
\caption{Using retrieved supervision examples for general domain adaptaion. The first number in each cell of the table is the score obtained by the corresponding model, which is followed by the change in the performance w.r.t domain-specific model or test sample-specific demonstrations.
}
\vspace{-0.2cm}
\label{tab:result3}
\end{table*}

\subsubsection*{Finding: Strategic adaptation is advantageous in very low data scenarios
}
Figure~\ref{fig:result2-codet5} and~\ref{fig:result2-codex} demonstrate the performance of the CodeT5 and Codex models. 
For CodeT5, it contains the performance obtained without adaptation (0-shot), as well as after in-domain few-shot fine-tuning~(additional results for LoRA are presented in Appendix Section~\ref{app:metrics}). 
None of the evaluated methods perform comparably in zero-shot setting to those with few-shot domain adaptation - whether on examples retrieved from training data or obtained from test domains. So these training methods do not result in a general-purpose model that handles out-of-domain generalization well. 

The same pattern is evident in the challenge evaluation scenario, presented in Figure~\ref{fig:result2-codet5-vault}. From this figure, we also conclude that retrieved supervision is less effective when supervised and test examples are extracted from different datasets - even when both are collected from the same source, i.e. GitHub. While we have done our best to process the data in The Vault dataset as similar to the processing done in CodeSearchNet, there must still be subtle differences remaining from data collection, once again emphasizing how sensitive code models are even to minute changes.

Adapting the model trained with MTL objective to test domains with the help of stratified supervision provides a considerable boost to the performance of CodeT5 and Codex. 
Results for CodeT5 are shown in Figure~\ref{fig:result2-var-ret} with bars marked ``ret $k$'', where $k$ refers to the number of examples included in $\tau_{ret}$ per test example. 
Figure~\ref{fig:result2-codex} reports Codex performance with 4 or 8 retrieved demonstrations as ``ICL ret 4'' and ``ICL ret 8'' respectively. 

First of all, we notice that there is a saturation in terms of gained performance vs the number of stratified supervision or demonstration examples used. 
For CodeT5 using 32 examples per test instance is almost always worse than using 4 or 8 examples. 
For Codex, using 4 or 8 examples results in approximately the same performance.

Next, for code summarization, retrieving 4 or 8 examples from out-of-domain train data leads to performance comparable, or even better, than that of the model adapted using 8 examples from the test domain. This trend is observed for both Codex and CodeT5, particularly strongly when generalizing to new repositories and new organizations.  
A similar trend can be observed for code generation, and to a much stronger degree for CodeT5 - stratified supervision models can even outperform models trained with 32 examples from the test domain.
While the performance of the stratified supervision models plateau after a certain number of examples, supervision on in-domain samples does not demonstrate such a trend.

\begin{question}
Can we have more generic solutions for out-of-domain generalization?
\end{question}
From our analysis of RQ2, we see that models can generalize better to new domains without relying on labeled data from that domain.
Unfortunately, this still requires adapting to every test domain individually for CodeT5, and even more strictly - to every test sample individually - for Codex. 
For example, for CodeT5, this means maintaining multiple copies of the model, performing the training for the adaptation stage multiple times, and storing a large amount of out-of-domain data to retrieve examples from.  
In this RQ, we experiment with approaches that would eliminate the need to train CodeT5 on multiple domains separately. 
For Codex, we experiment with sampling from demonstrations collected for the entire domain.  
For CodeT5, we try two approaches. First, we finetune it on the \textit{combined} set of $\tau_{ret}$ for all domains. We also try using \textit{fast vote-k} algorithm~\citep{DBLP:journals/corr/abs-2209-01975}, which selects representative examples from the supervision dataset, while ensuring diversity among selected examples. 
For Codex, for a query from $\tau_{test}$, we consider sampling 4 or 8 demonstration examples from $\tau_{ret}$.
\vspace{-0.1cm}
\subsubsection*{Finding: Multi-domain code generation models do not require a large performance sacrifice.}
The results for both models are presented in Table~\ref{tab:result3}. Results for CodeT5 for this experiment are referred to as ``FT: combined $k$'', where $k$ is the number of retrieved examples per test example. Fast vote-k is less effective as an adaptation technique compared to fine-tuning on a combined set of retrieved examples, and the results for it are presented in the Appendix Section~\ref{app:similarity-sa}. 
As can be seen, training a single model on combined retrieved samples results in a moderate drop in performance for code summarization, and a negligible drop for code generation. 
In other words, a model finetuned on stratified supervision data for new domains can be a viable solution for the out-of-domain generalization problem for code generation.
Interestingly, this also indicates that for code generation, good performance on one domain does not hinder the performance on another domain, i.e. there is little to no negative transfer between different domains.

For Codex, the results of the experiment are referred to as ``ICL: $k \text{ from } \tau_{ret}$'' in Table~\ref{tab:result3}, where $k$ is the number of sampled demonstrations. 
It appears that for Codex replacing demonstrations selected for individual examples with those selected for a domain introduce too much noise, and degrade the performance a lot because of the high sensitivity of ICL to demonstrations. 

\vspace{-0.2cm}
\section{Conclusion}
\vspace{-0.2cm}
We evaluate large language models for code - CodeT5, Codex (code-cushman-001), and ChatGPT (gpt-3.5-turbo) - on two fundamental code applications - code generation and code summarization. 
We study how the models perform under distribution shifts that can commonly occur due to the nature of the software. 
We experiment with three granularities for defining domains in applications for code - organization, project, and module or folder. 
Our experiments show that all models evaluated are susceptible to reduced performance due to domain shifts. 
We experiment with a number of training and domain adaptation techniques for achieving better out-of-domain generalization. 
We discover that retrieving similar out-of-domain examples from training data is the most effective approach for adapting to new domains in the absence of in-domain data.
In addition, we experiment with adapting models to multiple new domains simultaneously and find that such models can perform very well for code generation. 
However, we find the generality of the model to be a tradeoff for its performance for code summarization.

\newpage
\vspace{-0.2cm}
\section{Limitations and Threats to Validity}
\vspace{-0.2cm}
As can be seen from Table~\ref{tab:dataset}, as a result of the process of filtering, we skew the data towards larger projects and eliminate from the dataset many samples that could potentially come from smaller projects. 
We believe that this step is necessary to make the results more reliable, due to the high variance that can be observed in datasets with very small test sets. 
However, we want to draw attention to this circumstance once more, to make sure that our findings are interpreted correctly.

\vspace{-0.2cm}
\section{Acknowledgements}
This research is supported in part by the Office of the Director of National Intelligence (ODNI), Intelligence Advanced Research Projects Activity (IARPA), via the HIATUS Program contract \#2022-22072200006, the DARPA MCS program under Contract No. N660011924033, the Defense Advanced Research Projects Agency with award W911NF-19-20271, NSF IIS 2048211, and gift awards from Google and Amazon. The views and conclusions contained herein are those of the authors and should not be interpreted as necessarily representing the official policies, either expressed or implied, of ODNI, IARPA, or the U.S. Government. We would like to thank all the collaborators in USC INK research lab for their constructive feedback on the work. 
S.A. was employed as a student researcher at Microsoft while working on this research.
\vspace{-0.2cm}

\bibliographystyle{acl_natbib.bst}
\bibliography{bibliography}
\clearpage
\newpage
\section{Appendix}
\subsection{Javascript Keywords}\label{app:js-kwords}
The Javascript keywords that we included in the CodeBleu implementation for evaluation is listed in table \ref{app:keywords}.
\label{app:keywords}
\begin{table*}[t]
\centering
\begin{tabular}{|l|p{0.8\textwidth}|}
\hline
\multicolumn{1}{|c|}{\textit{\textbf{Languages}}} & \multicolumn{1}{c|}{\textit{\textbf{Keywords}}}                                                   \tabularnewline \hline
JavaScript                                             & await, break, case, catch, class, const, continue, debugger, default, delete, do, else, enum, export, extends, false, finally, for, function, if, implements, import, in, instanceof, interface, let, new, null, package, private, protected, public, return, super, switch, static, this, throw, try, true, typeof, var, void, while, with, yield 
\tabularnewline \hline
\end{tabular}
\caption{Keywords used for CodeBLEU evaluation}
\label{table:keywords}   
\end{table*}

\begin{table*}[t]
\centering
\resizebox{\textwidth}{!}{%
\begin{tabular}{lrrrrrrrrr}
\toprule
\multicolumn{1}{c}{
    \multirow{2}{*}{\textbf{Code generation}}
} &
\multicolumn{3}{c}{\textbf{folder}} & 
\multicolumn{3}{c}{\textbf{repo}} & 
\multicolumn{3}{c}{\textbf{org}} \\
\cmidrule(lr){2-4}
\cmidrule(lr){5-7}
\cmidrule(lr){8-10}
\multicolumn{1}{c}{} & 
\multicolumn{1}{c}{\textbf{8-shot}} & 
\multicolumn{1}{c}{\textbf{16-shot}} & 
\multicolumn{1}{c}{\textbf{32-shot}} & 
\multicolumn{1}{c}{\textbf{8-shot}} & 
\multicolumn{1}{c}{\textbf{16-shot}} & 
\multicolumn{1}{c}{\textbf{32-shot}} & 
\multicolumn{1}{c}{\textbf{8-shot}} & 
\multicolumn{1}{c}{\textbf{16-shot}} & 
\multicolumn{1}{c}{\textbf{32-shot}} \\
\midrule
CodeT5 FT ID 
& 19.36 &	20.92 & 21.95
& 20.42 &	22.44 &	24.47
& 19.29	& 20.73	& 22.6  \\
CodeT5 LoRA ID 
& 20.05 &	21.66 &	22.56
& 20.81 &	23.12 &	24.52
& 20.08	& 21.28	& 22.99\\
CodeT5 FT random 
& 17.61	& 18.03 &	17.94
& 16.92	& 17.50 & 17.59
& 16.47	& 17.46	& 17.85 \\
CodeT5 LoRA random 
& 17.87	& 18.02 &	17.81
& 17.45	& 17.15 &	17.63
& 17.24	& 17.13 &	17.29\\
\midrule
Codex ICL ID
& 28.78 & - & - 
& 31.05 & - & - 
& 29.19 & - & -  \\
Codex ICL random
& 20.62 & - & - 
& 20.87 & - & - 
& 21.10 & - & - \\
Codex instr. only \textit{(0-shot)}
& (10.24) & - & - 
& (10.60) & - & - 
& (10.25) & - & - \\
\bottomrule          
\end{tabular}
}
\caption{Comparison of model performance for code generation on in-domain (\textbf{ID}) vs out-of-domain (\textbf{random}) test data. Reported metric is ChrF (higher is better).
}
\label{tab:text-to-code-chrf}
\end{table*}

\begin{table*}[t]
\centering
\resizebox{\textwidth}{!}{%
\begin{tabular}{lrrrrrrrrr}
\toprule
\multicolumn{1}{c}{
    \multirow{2}{*}{\textbf{Code generation}}
} &
\multicolumn{3}{c}{\textbf{folder}} & 
\multicolumn{3}{c}{\textbf{repo}} & 
\multicolumn{3}{c}{\textbf{org}} \\
\cmidrule(lr){2-4}
\cmidrule(lr){5-7}
\cmidrule(lr){8-10}
\multicolumn{1}{c}{} & 
\multicolumn{1}{c}{\textbf{8-shot}} & 
\multicolumn{1}{c}{\textbf{16-shot}} & 
\multicolumn{1}{c}{\textbf{32-shot}} & 
\multicolumn{1}{c}{\textbf{8-shot}} & 
\multicolumn{1}{c}{\textbf{16-shot}} & 
\multicolumn{1}{c}{\textbf{32-shot}} & 
\multicolumn{1}{c}{\textbf{8-shot}} & 
\multicolumn{1}{c}{\textbf{16-shot}} & 
\multicolumn{1}{c}{\textbf{32-shot}} \\
\midrule
CodeT5 FT ID 
& 14.15 &	15.84 & 16.73
& 14.93	& 16.98	& 19.19
& 13.75	& 14.93	& 16.94  \\
CodeT5 LoRA ID 
& 14.49	& 16.58	& 17.87
& 15.47	& 17.69	& 19.60
& 14.10	& 15.48	& 17.61 \\
CodeT5 FT random 
& 11.34	& 11.62	& 11.73
& 9.91	& 10.10	& 10.32
& 9.49	& 10.20	& 10.68 \\
CodeT5 LoRA random 
& 11.45	& 12.05	& 12.58
& 10.09	& 10.04	& 11.08
& 10.15	& 10.30	& 11.15\\
\midrule
Codex ICL ID
& 23.70 & - & - 
& 24.62 & - & - 
& 22.58 & - & -  \\
Codex ICL random
& 15.76 & - & - 
& 15.67 & - & - 
& 15.81 & - & - \\
Codex instr. only \textit{(0-shot)}
& (6.44) & - & - 
& (6.50) & - & - 
& (6.18) & - & - \\
\bottomrule          
\end{tabular}
}
\caption{Comparison of model performance for code generation on in-domain (\textbf{ID}) vs out-of-domain (\textbf{random}) test data. Reported metric is RougeL (higher is better).
}
\label{tab:text-to-code-rougel}
\end{table*}

\begin{table*}[t]
\centering
\resizebox{\textwidth}{!}{%
\begin{tabular}{lrrrrrrrrr}
\toprule
\multicolumn{1}{c}{
    \multirow{2}{*}{\textbf{Code generation}}
} &
\multicolumn{3}{c}{\textbf{folder}} & 
\multicolumn{3}{c}{\textbf{repo}} & 
\multicolumn{3}{c}{\textbf{org}} \\
\cmidrule(lr){2-4}
\cmidrule(lr){5-7}
\cmidrule(lr){8-10}
\multicolumn{1}{c}{} & 
\multicolumn{1}{c}{\textbf{8-shot}} & 
\multicolumn{1}{c}{\textbf{16-shot}} & 
\multicolumn{1}{c}{\textbf{32-shot}} & 
\multicolumn{1}{c}{\textbf{8-shot}} & 
\multicolumn{1}{c}{\textbf{16-shot}} & 
\multicolumn{1}{c}{\textbf{32-shot}} & 
\multicolumn{1}{c}{\textbf{8-shot}} & 
\multicolumn{1}{c}{\textbf{16-shot}} & 
\multicolumn{1}{c}{\textbf{32-shot}} \\
\midrule
CodeT5 FT ID 
& 0.68 / 0.68	& 0.69 / 0.68 & 0.69 / 0.69
& 0.69 / 0.69&	0.69 / 0.68&	0.68 / 0.67
& 0.69 / 0.67	&0.69 / 0.68	&0.70 / 0.69  \\
CodeT5 LoRA ID 
& 0.68 / 0.67	& 0.69 / 0.68	& 0.70 / 0.69
& 0.69 / 0.68	&0.70 / 0.70	&0.71 / 0.71
& 0.69 / 0.68	&0.69 / 0.68	&0.71 / 0.69 \\
CodeT5 FT random 
& 0.65 / 0.66	&0.66 / 0.66	&0.66 / 0.66
& 0.66 / 0.65	&0.66 / 0.66	&0.66 / 0.66
& 0.65 / 0.65	&0.65 / 0.65	&0.65 / 0.65\\
CodeT5 LoRA random 
& 0.65 / 0.65 &0.65 / 0.65	&0.66 / 0.66
& 0.66 / 0.66	&0.65 / 0.65	&0.66 / 0.66
& 0.65 / 0.65	&0.65 / 0.65	&0.66 / 0.66\\
\midrule
Codex ICL ID
& 0.74 / 0.72 & - & - 
& 0.75 / 0.73 & - & - 
& 0.74 / 0.72 & - & -  \\
Codex ICL random
& 0.69 / 0.67 & - & - 
& 0.70 / 0.68 & - & - 
& 0.69 / 0.67 & - & - \\
Codex instr. only \textit{(0-shot)}
& 0.62 / 0.61 & - & - 
& 0.63 / 0.62 & - & - 
& 0.63 / 0.62 & - & - \\
\bottomrule          
\end{tabular}
}
\caption{Comparison of model performance for code generation on in-domain (\textbf{ID}) vs out-of-domain (\textbf{random}) test data. Reported metric in each cell is CodeBERTScore F1 on the left (higher is better), and CodeBERTScore F3 on the right (higher is better).
}
\label{tab:text-to-code-codebertscore}
\end{table*}

\subsection{Extended Background}\label{app:extended background}

\subsubsection{Meta-learning and Multi-task-learning}
\paragraph{Meta-learning} focuses on adapting knowledge gained from previous tasks to be applied to new tasks with limited training examples. Most meta-learning algorithms can be categorized into three groups: 1) Black-box meta-learning approaches~\citep{Santoro2016MetaLearningWM} train a black-box model to take in training data of a target task to output parameters for the neural network used for making prediction for that task; 2) Optimization-based methods~\citep{Finn2017ModelAgnosticMF, Finn2018ProbabilisticMM, Antoniou2018HowTT} uses gradient descent to learn model parameters which can be adapted to a future target task with few gradient steps on a few-shot training dataset; 3) Non-parametric methods~\citep{vinyals2017matching, Snell2017PrototypicalNF, Sung2017LearningTC, Koch2015SiameseNN} learns a metric space in which predictions can be performed by computing some similarity metric, like distance and cosine similarity, to representations of each class. In our work, we are using the MAML~\citep{Finn2017ModelAgnosticMF} approach, which is a gradient-based method and learns model initialization (i.e., initial parameters) that is amenable to fast fine-tuning with few instances. This method is a conceptually simple and model-agnostic algorithm that has been shown to outperform existing approaches in several tasks.
\paragraph{Multi-task Learning} aims to jointly learn several related tasks providing a generalized representation with the added benefit of 
 compute and memory in terms of shared model parameters~\citep{Yang2016DeepMR, Caruana1997MultitaskL, Meyerson2019ModularUR}. MTL also has a regularization effect on the model parameters.  By definition, MTL aims to solve a fixed number of known tasks, whereas the point of meta-learning is often to solve unseen future tasks. But both methods capture a good prior from the training tasks, which can be used for getting model parameters for future target tasks.

In our work, we have experimented with both MAML and multi-task learning to check which of the method gives us a better prior for few-shot performance in our setting.
\subsubsection{Few-shot Methods}
\paragraph{Parameter-efficient finetuning:}Conventional fine-tuning methods retrains all the model parameters for every new task, which becomes infeasible as the model size increases to the level of GPT-3. In recent times, parameter-efficient methods have been studied and it has been demonstrated that state-of-the-art PEFT methods can match the performance of finetuning all the model's parameters while updating only a tiny fraction of the model parameters. Initially adapters~\citep{Raffel2019ExploringTL, Houlsby2019ParameterEfficientTL, Bapna2019SimpleSA} were introduced, which are new feed-forward modules added between the layers of the fixed pre-trained model. Since then, various sophisticated PEFT methods have been proposed, including methods like LoRA that produce low-rank updates~\citep{Hu2021LoRALA} and prompt tuning ~\citep{Lester2021ThePO} and prefix-tuning ~\citep{Li2021PrefixTuningOC} concatenate learned continuous embeddings to the model’s input or activations to induce it to perform a task. 

\paragraph{Retrieval-based Example selection:} In a study conducted by \citet{Liu2021WhatMG} , they explored how different prompts can impact the performance of GPT-3 and found that the use of in-context examples has a significant influence on the downstream results. To achieve this, they utilized an unsupervised sentence encoder to encode training examples and then retrieved the nearest neighbors for each test instance. On a similar note, \citet{Das2021CasebasedRF} developed a supervised prompt retriever for answering knowledge-based questions. Their method used tailored supervision specifically designed for knowledge-based queries and relied on surface similarity between formal queries. Furthermore,  \citet{Shin2021ConstrainedLM} employed GPT-3 to select examples for the prompt in few-shot semantic parsing. They demonstrated the effectiveness of this approach by using GPT-3 to identify relevant examples for the prompt, which in turn improved the overall performance of the system.

\begin{figure}[h]
     \centering
     \includegraphics[trim=0 0 0 0,clip,width=0.35\textwidth]{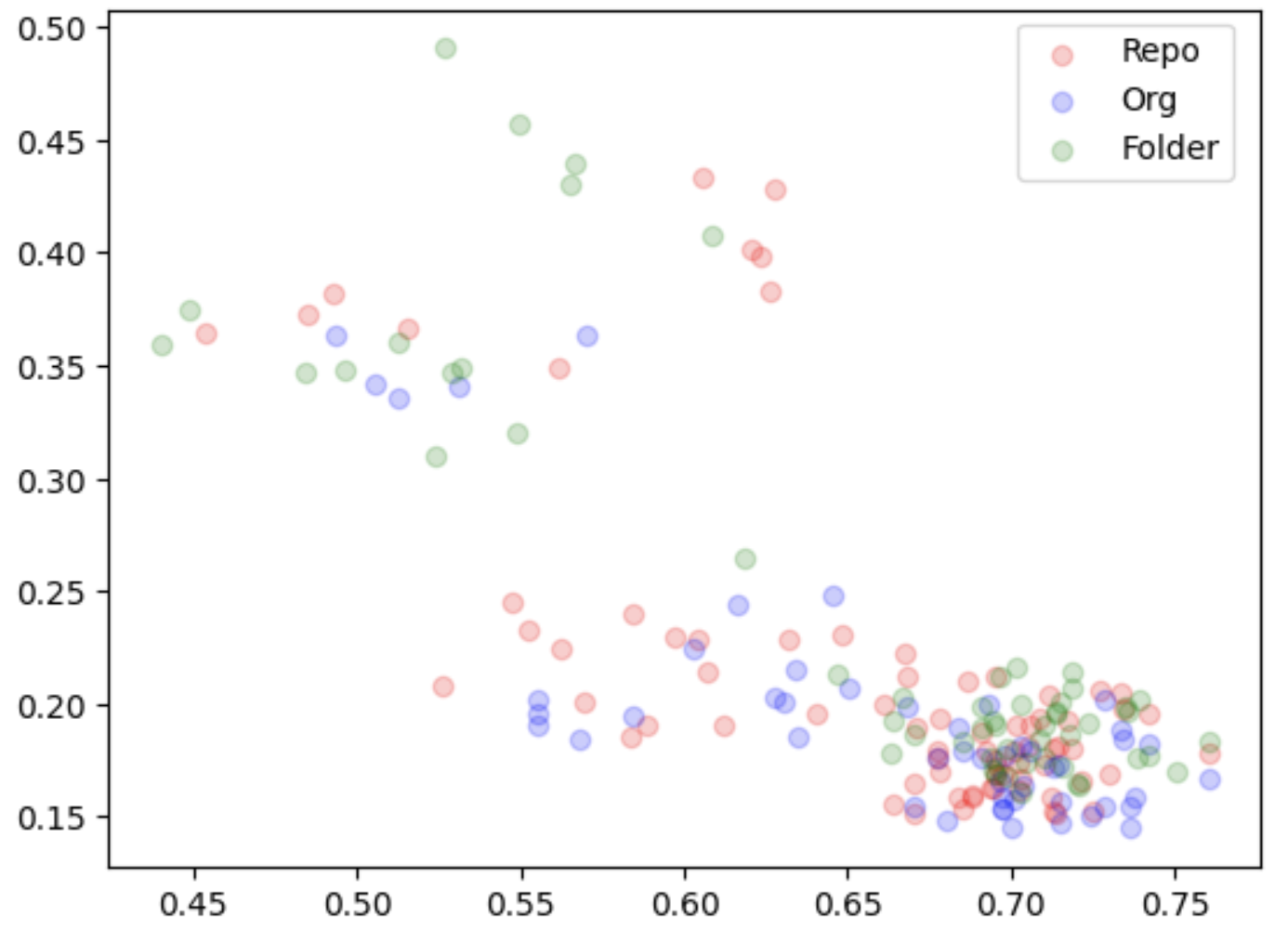}
     \caption{Each dot signifies a domain. Average pairwise similarities of examples within each domain (x axis) plotted against average similarities of that domain to all other domains (y axis). }
     \label{fig:domain-viz}
\end{figure}

\subsection{Domain split visualization}\label{app:data-viz}
To better understand how different splits of domains are different from each other, we visualize our resulting test domains in Figure~\ref{fig:domain-viz}. We plot each domain as a dot, where different colors correspond to different splits. X axis demonstrates average pairwise similarity of examples within a domain, i.e. x coordinate of a domain corresponds to how uniform examples within a domain are. Y axis demonstrates pairwise similarities of examples within a domain to examples in all other domains, i.e. y coordinate of a domain demonstrates its similarity to other domains. From the figure we see that the vast majority of domains are clustered in the lower right corner, which corresponds to the domains that are uniform, and dissimilar to other domains. A small handful of domains are located in the upper left corner, that corresponds to domains with dissimilar examples within itself, but higher similarity to other domains. It is notable, that quantitatively, upper left corner contains more folders than repos, and more repos than orgs. We hypothesize, that such distribution could be explained by functional, rather than hierarchicals similarities across domains. A clear example of such instance can be a folder with utility functions that can have high similarity to other folders with utility functions, all the while individual functions within that folder are implementing different utilities, and thus - are dissimilar.  

\subsection{Models}\label{app:model}
\paragraph{CodeT5:} CodeT5~\citep{Wang2021CodeT5IU} is a pretrained encoder-decoder transformer model based on T5~\citep{Raffel2019ExploringTL} for programming languages. It uses a unified framework to support code understanding and generation tasks seamlessly. To improve the model's ability to handle the unique characteristics of programming languages, CodeT5 is trained on an identifier-aware pretraining task. Additionally, the model is trained to exploit  user-written code comments with a bimodal dual-generation task for better alignment between natural language and programming languages. This makes this model suitable for the applications that we consider. For both of our applications, we used the CodeT5-large model~\citep{DBLP:journals/corr/abs-2207-01780} without making any changes to the model architecture.

\paragraph{Codex} Codex~\citep{Chen2021EvaluatingLL} is the language model for code released by OpenAI. It is a GPT language model finetuned on 54 million public software repositories hosted on GitHub, containing 179 GB of unique Python files under 1 MB. VLLMs are capable of zero-shot generalization to unseen tasks, which is achieved by providing them with an \textit{instruction} of what the model is expected to do. This allowed us to successfully evaluate Codex for both code generation and code summarization without any need for training. 

\paragraph{ChatGPT} ChatGPT is a conversational variant derived from InstructGPT/GPT 3.5 model~\citep{Ouyang2022TrainingLM}. It features a dialogue interface and is trained using a more refined objective function called Reinforcement Learning from Human Feedback (RLHF)~\citep{Christiano2017DeepRL}. However, there is currently limited information available regarding the specific architecture and training data employed in the creation of ChatGPT. We utilize the GPT-3.5 Turbo API, provided by OpenAI, to access ChatGPT for conducting our experiments. This API version allows a maximum token length restriction of 4096 tokens.

\subsection{Hyperparameters and training details}\label{app:hyperparams}
For full finetuning of CodeT5, we updated the model for 500 steps using batch size of 8, the best model was identified by the performance on the $\tau_{dev}$ portion. 
For LoRA, we use a rank of 4 with an initialization scale of 0.01 and update all the attention and feedforward layers. We train for 1000 steps with a batch size of 8.

For multitask learning (MTL) of CodeT5, we update the model for 150K steps on 80\% of the $X_{train}$ data, using a batch size of 4. The best checkpoint is selected by evaluating the model on the remaining 20\% of $X_{train}$ which was held-out from training.
For dual-gen MTL, we followed the same train/dev division strategy as for MTL for code generation, and updated the model for 150K steps with batch size of 4. 
The best checkpoints were again decided by evaluating the model on the created development set. 
In particular, we selected two checkpoints - one according to CodeBLEU metric, and another according to BLEU metric for code generation and code summarization respectively.
For Model-agnostic meta-learning, we updated the model from the pretrained CodeT5 checkpoint for 10K steps and used the last checkpoint in our experiments.

\subsection{The Vault} \label{app:vault}
The Vault is a multilingual dataset extracted from GitHub. Despite the fact that it comes pretokenized, we noticed that some of the preprocessing for The Vault is different from the preprocessing of CodeSearchNet. For example, while CodeSearchNet function body may have inlined comments, the Vault functions are stripped of those. On the other side, the Vault docstring typically includes function parameter documentation, whereas the CodeSearchNet omits those. On average, CodeSearchNet function docstrings are also shorter than those of the Vault. In our work, we processed the Vault dataset, to fix these inconsistencies and make new data points consistent with data from CodeSearchNet. 

\subsection{Additional experimental results}\label{app:metrics}

Besides the experiments presented in the main paper, in this section, we report some additional experiments. 
Tables~\ref{tab:text-to-code-chrf},~\ref{tab:text-to-code-rougel} and~\ref{tab:text-to-code-codebertscore} report results for code generation as measured using chrF, RougeL and CodeBERTScore metrics correspondingly.

\begin{figure*}[h]
     \centering
     \includegraphics[trim=0 0 0 0,clip,width=\textwidth]{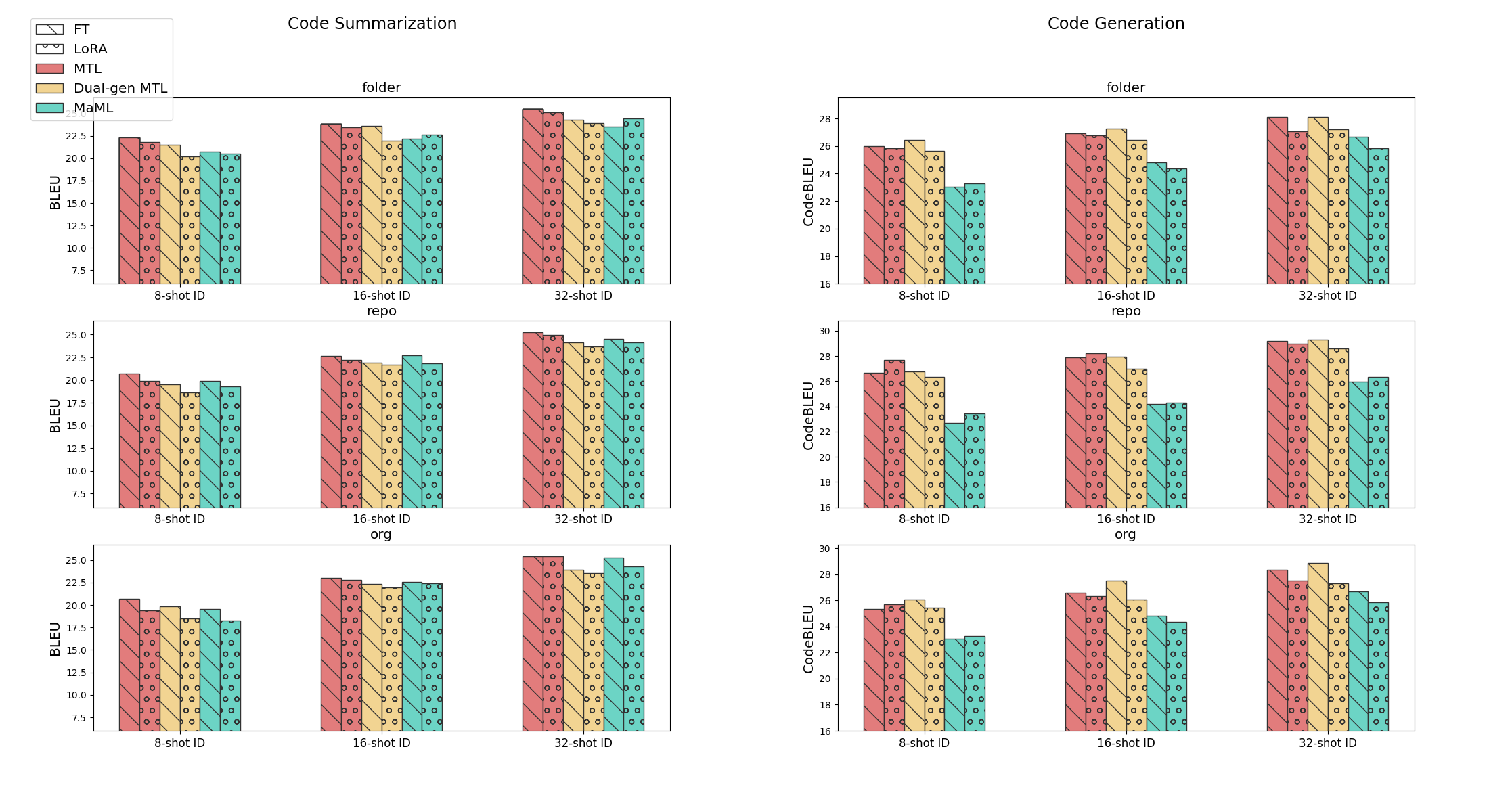}
     \caption{Performance for CodeT5 model finetuned with LoRA compared to regular finetuning.}
     \label{fig:result2-codet5-lora}
\end{figure*}

Additionally, Figure~\ref{fig:result2-codet5-lora} illustrates how LoRA parameter efficient finetuning method compares to the full model finetuning for CodeT5. 

\begin{table}[h]
\centering
\resizebox{0.48\textwidth}{!}{
\begin{tabular}{lrrrrrr}
\toprule
 & \multicolumn{3}{c}{\begin{tabular}[c]{@{}c@{}}Code Summarization\\ 
 BLEU \end{tabular}} & \multicolumn{3}{@{\hskip 0.6cm}c}{\begin{tabular}[c]{@{}c@{}}Code Generation\\ CodeBLEU \end{tabular}} \\
 \cmidrule(lr){2-4}
 \cmidrule(lr){5-7}
 & \multicolumn{1}{c}{org} & \multicolumn{1}{c}{repo} & \multicolumn{1}{c}{folder} & \multicolumn{1}{c}{org} & \multicolumn{1}{c}{repo} & \multicolumn{1}{c}{folder} \\
\midrule
IsoScore (4) 
& 16.71 & 16.57 & 15.47 & 15.05 & 16.01 & 14.93 \\
IsoScore (8) 
& 17.27 & 16.72 &	15.71 &	15.32 &	16.55 &	15.28 \\
IsoScore (32) 
& 17.46 &	16.90 &	14.34 &	16.13 &	17.89 &	16.26 \\
\bottomrule
\end{tabular}}
\caption{ Results for CodeT5 model using IsoScore for measuring embedding similarity and supervising with retrieved examples from train data.
}
\label{tab:result-isoscore}
\end{table}

\subsection{IsoScore}\label{app:similarity-isoscore}
IsoScore is a similarity metric of isotropy of an embedding space. The way we use it to measure similarity is by computing IsoScore value of a combined set of test example embeddings and every individual training set embedding. The ``closest'' examples selected for supervision are the ones that resulted in the largest IsoScore value for each set of test examples. We then use the same number of supervision examples as we used with cosine similarity - selecting 4*32, 8*32, or 32*32 ``closest'' examples for supervision. The results for model adapted using IsoScore metric similarity are reported in Table~\ref{tab:result-isoscore}.

\subsection{Fast vote-k}\label{app:similarity-sa}
To make the setup for fast vote-k similar to the version with the combination of nearest examples, we run this algorithm to select 4*32 (128), 8*32 (256), and 32*32 (1024) supervision examples. 
Table~\ref{tab:result-vote-k} show results obtained for a CodeT5 MTL model that has additionally been finetuned using a set of examples obtained from fast vote-k algorithm. 

\begin{table}[h]
\centering
\resizebox{0.48\textwidth}{!}{
\begin{tabular}{lrrrrrr}
\toprule
 & \multicolumn{3}{c}{\begin{tabular}[c]{@{}c@{}}Code Summarization\\ 
 BLEU \end{tabular}} & \multicolumn{3}{@{\hskip 0.6cm}c}{\begin{tabular}[c]{@{}c@{}}Code Generation\\ CodeBLEU \end{tabular}} \\
 \cmidrule(lr){2-4}
 \cmidrule(lr){5-7}
 & \multicolumn{1}{c}{org} & \multicolumn{1}{c}{repo} & \multicolumn{1}{c}{folder} & \multicolumn{1}{c}{org} & \multicolumn{1}{c}{repo} & \multicolumn{1}{c}{folder} \\
\midrule
Fast vote-k (4) 
& 10.96 &	12.34 &	10.33 &	24.96 &	25.76 &	24.77 \\
Fast vote-k (8) 
& 11.40 &	12.74 &	10.60 &	25.10 &	26.21 &	25.14 \\
Fast vote-k (32) 
& 10.84&	12.03 &	10.06 &	24.25 &	25.06 &	24.17 \\
\bottomrule
\end{tabular}}
\caption{ Results for CodeT5 model using Fast Vote-k for measuring embedding similarity and supervising with retrieved examples from train data.
}
\label{tab:result-vote-k}
\end{table}

\subsection{Instructions for Codex and ChatGPT} \label{app:instructions}

Table~\ref{tab:instructions} contains list of instructions we used with Codex and ChatGPT models in instruction-only and in-context learning scenarios. 

\begin{table*}[t]
\centering
\resizebox{\textwidth}{!}{
\begin{tabular}{lll}
\toprule
Copilot & Task instruction & \begin{tabular}[c]{@{}l@{}}"Write in javascript:",\\ "Write code:", \\ "Summarize code:",\\ "Summarize javascript snippet:",\\ "Write code intent:"\end{tabular}\\
 & \begin{tabular}[c]{@{}l@{}}Demonstration example \\ template\end{tabular} & \begin{tabular}[c]{@{}l@{}}"Intent: \{text\} \textbackslash{}\textbackslash{}n Snippet: \{code\}\textbackslash{}\textbackslash{}n\textbackslash{}\textbackslash{}n", \\ "Intent: \{text\} \textbackslash{}\textbackslash{}n Code: \{code\}\textbackslash{}\textbackslash{}n\textbackslash{}\textbackslash{}n",\\ "Code: \{code\} \textbackslash{}\textbackslash{}n Intent: \{text\}\textbackslash{}\textbackslash{}n\textbackslash{}\textbackslash{}n",\\ "Code: \{code\} \textbackslash{}\textbackslash{}n Summary: \{text\}\textbackslash{}\textbackslash{}n\textbackslash{}\textbackslash{}n",\\ "Snippet: \{code\} \textbackslash{}\textbackslash{}n Intent: \{text\}\textbackslash{}\textbackslash{}n\textbackslash{}\textbackslash{}n",\\ "Snippet: \{code\} \textbackslash{}\textbackslash{}n Summary: \{text\}\textbackslash{}\textbackslash{}n\textbackslash{}\textbackslash{}n",\end{tabular} \\
\midrule
ChatGPT & System messages & \begin{tabular}[c]{@{}l@{}}'You are a helpful assistant that writes JavaScript code based on English description. \\ You only output code without any English text.'\\ ``You are a helpful assistant that writes single sentence summarizes for JavaScript code in English. \\ You only output code summary without any other English text.''\end{tabular} \\
 & Task instruction & \begin{tabular}[c]{@{}l@{}}"Write a single sentence summary for the following JavaScript code in English. "\\ "Implement this functionality using JavaScript. "\end{tabular} \\
 & \begin{tabular}[c]{@{}l@{}}Demonstration example \\ template\end{tabular} & \begin{tabular}[c]{@{}l@{}}{[}"Below are some examples of JavaScript code implemented based on English summary. \textbackslash{}n",\\ "Summary: \{text\}\textbackslash{}nCode: \{code\}\textbackslash{}n\textbackslash{}n"{]}\\ {[}"Below are some examples of English summaries of JavaScript code. \textbackslash{}n",\\ "Code: \{code\}\textbackslash{}nSummary: \{text\}\textbackslash{}n\textbackslash{}n"{]}\end{tabular} \\
 \bottomrule
\end{tabular}}
\caption{Task instructions and demonstration templates used for generating results in the experiments with Codex and ChatGPT.}
\label{tab:instructions}
\end{table*}

\subsection{Sample outputs}
Table~\ref{tab:sample-outputs} presents some examples and the outputs obtained by different models for those. Here we can see that CodeT5 model finetuned on in-domain examples sometimes has the advantage of having relevant context and thus is using correct member names as opposed to other models. On the other hand, we also see that similar out-of-domain examples from the train split can in fact be near duplicates of the ones in the test split. As a result, the model supervised with retrieved examples may generate output that is extremely close to that of the gold test data.

\begin{table*}[]
\resizebox{\textwidth}{!}{
\begin{tabular}{llllll}
\toprule
\multicolumn{1}{c}{\textbf{Input}} & \multicolumn{1}{c}{\textbf{Gold}} & \multicolumn{1}{c}{\textbf{\begin{tabular}[c]{@{}c@{}}CodeT5\\ MTL (0-shot)\end{tabular}}} & \multicolumn{1}{c}{\textbf{\begin{tabular}[c]{@{}c@{}}CodeT5 \\ MTL + ID (32-shot)\end{tabular}}} & \multicolumn{1}{c}{\textbf{\begin{tabular}[c]{@{}c@{}}CodeT5 \\ MTL + ret 4\end{tabular}}} & \multicolumn{1}{c}{\textbf{ChatGPT}} \\
\midrule
\raisebox{-4\height}{\begin{tabular}{@{}c@{}}Dispatch stack information \\  to all handlers\end{tabular}} & \raisebox{-\height}{\includegraphics[width=0.3\textwidth]{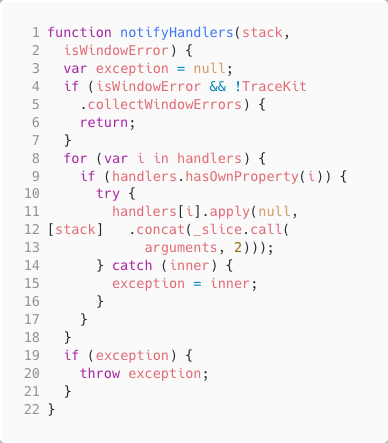}}&  
\raisebox{-\height}{\includegraphics[width=0.3\textwidth]{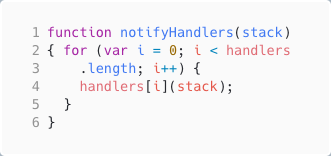}}&   \raisebox{-\height}{\includegraphics[width=0.3\textwidth]{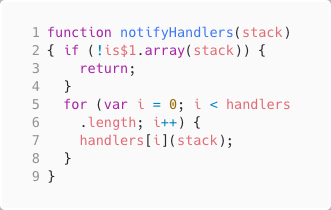}}&  \raisebox{-\height}{\includegraphics[width=0.3\textwidth]{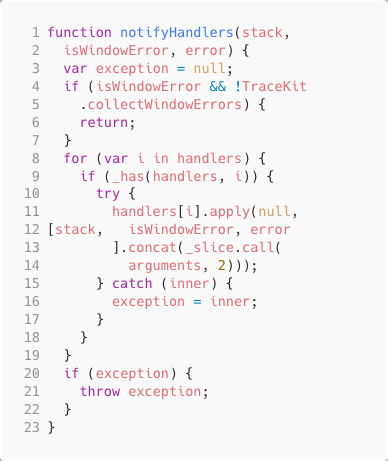}}&  \raisebox{-\height}{\includegraphics[width=0.3\textwidth]{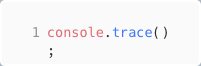}} \\
\midrule
\raisebox{-8\height}{\begin{tabular}{@{}c@{}}Setup captions\end{tabular}} & \raisebox{-\height}{\includegraphics[width=0.3\textwidth]{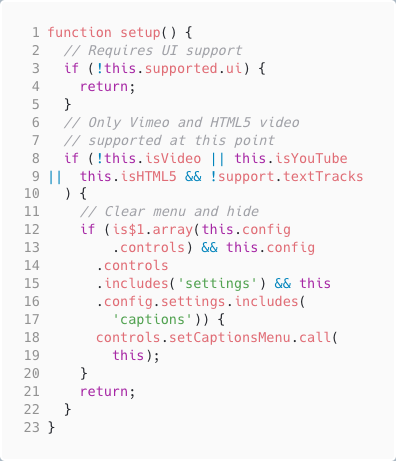}}&  
\raisebox{-\height}{\includegraphics[width=0.3\textwidth]{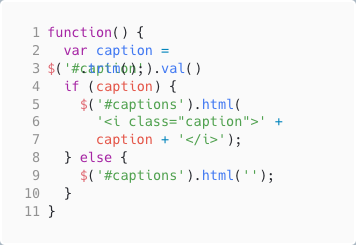}}&   \raisebox{-\height}{\includegraphics[width=0.3\textwidth]{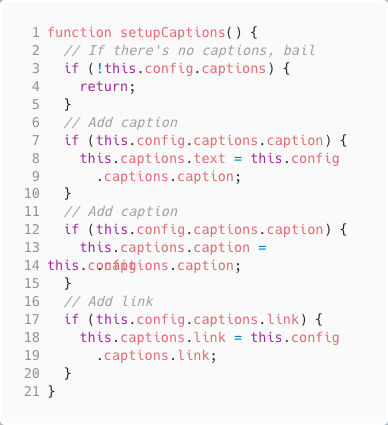}}&  \raisebox{-\height}{\includegraphics[width=0.3\textwidth]{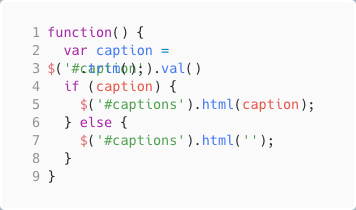}}&  \raisebox{-\height}{\includegraphics[width=0.3\textwidth]{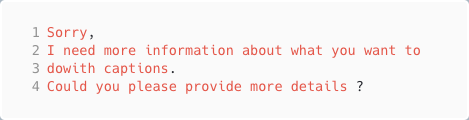}} \\
\midrule
\raisebox{-6\height}{\begin{tabular}{@{}c@{}}Toggle event listener\end{tabular}} & \raisebox{-\height}{\includegraphics[width=0.3\textwidth]{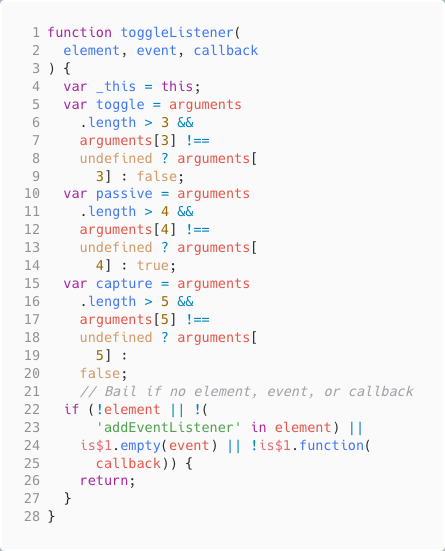}}&  
\raisebox{-\height}{\includegraphics[width=0.3\textwidth]{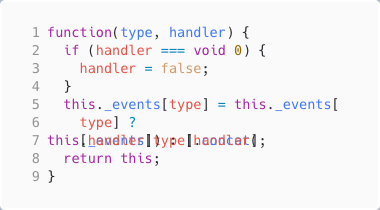}}&   \raisebox{-\height}{\includegraphics[width=0.3\textwidth]{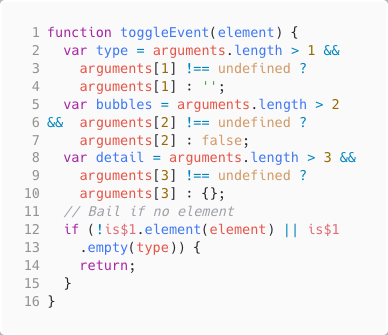}}&  \raisebox{-\height}{\includegraphics[width=0.3\textwidth]{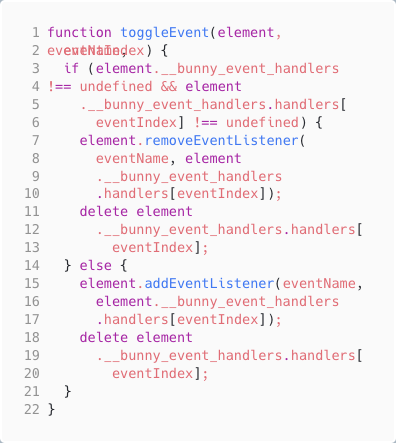}}&  \raisebox{-\height}{\includegraphics[width=0.3\textwidth]{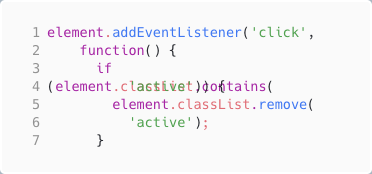}} \\
\midrule
\raisebox{-2\height}{\begin{tabular}{@{}c@{}}Returns the absolute path \\  to the class file\end{tabular}} & \raisebox{-\height}{\includegraphics[width=0.3\textwidth]{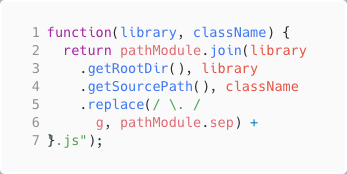}}&  
\raisebox{-\height}{\includegraphics[width=0.3\textwidth]{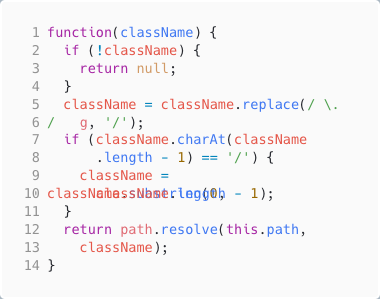}}&   \raisebox{-\height}{\includegraphics[width=0.3\textwidth]{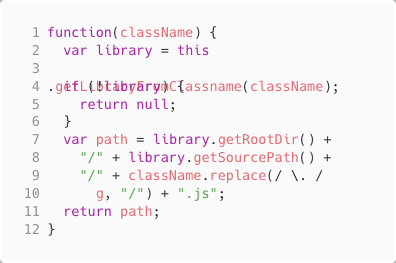}}&  \raisebox{-\height}{\includegraphics[width=0.3\textwidth]{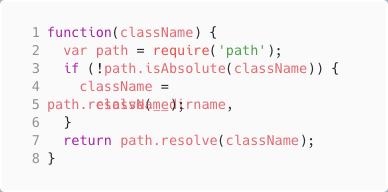}}&  \raisebox{-\height}{\includegraphics[width=0.3\textwidth]{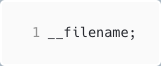}} \\
\midrule
\raisebox{-2\height}{\begin{tabular}{@{}c@{}}Returns the tag name of the\\   given library in the given contrib \\  repository if installed. \\ Returns false if not installed. \end{tabular}} & \raisebox{-\height}{\includegraphics[width=0.3\textwidth]{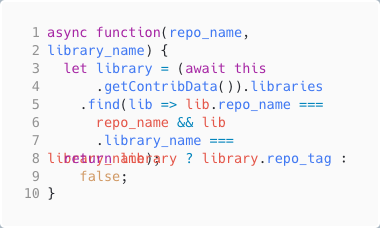}}&  
\raisebox{-\height}{\includegraphics[width=0.3\textwidth]{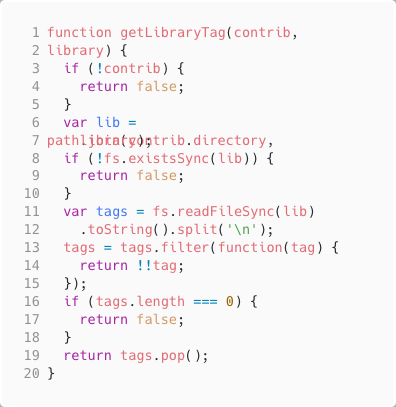}}&   \raisebox{-\height}{\includegraphics[width=0.3\textwidth]{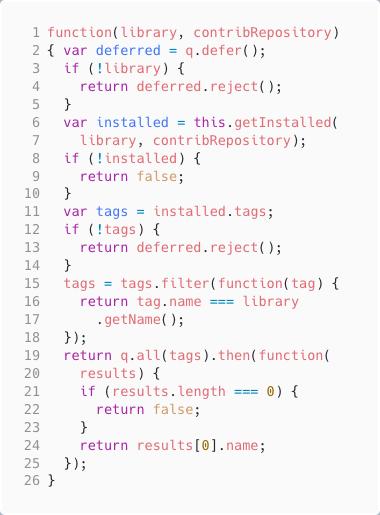}}&  \raisebox{-\height}{\includegraphics[width=0.3\textwidth]{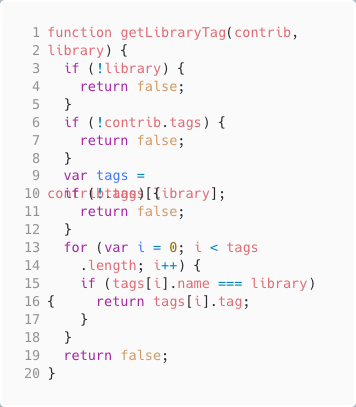}}&  \raisebox{-\height}{\includegraphics[width=0.3\textwidth]{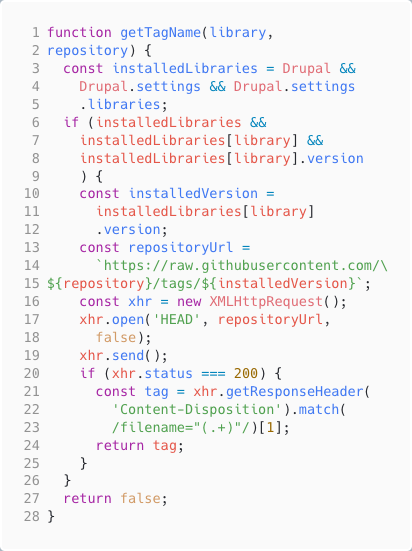}} \\
\bottomrule
\end{tabular}
}
\caption{Sample outputs from different models.}
\label{tab:sample-outputs}
\end{table*}

\end{document}